\pdfoutput=1

\documentclass[11pt]{article}

\usepackage{emnlp2021}

\usepackage{times}
\usepackage{latexsym}

\usepackage[T1]{fontenc}

\usepackage[utf8]{inputenc}

\usepackage{microtype}

\usepackage{amsfonts}
\usepackage{amsmath}
\usepackage{graphicx}
\usepackage{multirow}
\usepackage{subfigure}
\usepackage{enumitem}
\usepackage{float}
\usepackage{caption}
\usepackage{makecell}
\usepackage{amssymb}
\usepackage{booktabs}
\usepackage{cleveref}
\usepackage{arydshln}
\usepackage{pifont}
\usepackage{mathabx}
\newcommand{\cmark}{\ding{51}}%

\newcommand{\sub}{\textsubscript}

\usepackage{environ}
\NewEnviron{gather+}[1][1]{%
  \begin{equation*}
  \scalebox{#1}{$\begin{gathered}\BODY\end{gathered}$}
  \end{equation*}
}

\usepackage[russian,main=english]{babel} 

\usepackage{amsmath,amsfonts,bm}

\def\eqref#1{equation~\ref{#1}}

\def\1{\bm{1}}

\def\vb{{\bm{b}}}
\def\vc{{\bm{c}}}

\def\ve{{\bm{e}}}

\def\vn{{\bm{n}}}

\def\vs{{\bm{s}}}

\def\vy{{\bm{y}}}

\def\mE{{\bm{E}}}

\def\mK{{\bm{K}}}

\def\mQ{{\bm{Q}}}

\def\mV{{\bm{V}}}
\def\mW{{\bm{W}}}
\def\mX{{\bm{X}}}

\DeclareMathAlphabet{\mathsfit}{\encodingdefault}{\sfdefault}{m}{sl}
\SetMathAlphabet{\mathsfit}{bold}{\encodingdefault}{\sfdefault}{bx}{n}

\def\sR{{\mathbb{R}}}

\newcommand{\softmax}{\mathrm{softmax}}

\newcommand{\normltwo}{L^2}

\usepackage{xparse}
\NewDocumentCommand{\tong}{ mO{} }{\textcolor{orange}{\textsuperscript{\textit{Tong}}\textsf{\textbf{\small[#1]}}}}

\title{Char2Subword: Extending the Subword Embedding Space Using\\ Robust Character Compositionality}

\author{
Gustavo Aguilar,$^{\ddagger*}$
Bryan McCann,$^{\dagger**}$
Tong Niu,$^\dagger$ \\
\textbf{Nazneen Rajani,$^\dagger$} 
\textbf{Nitish Keskar,$^\dagger$} 
\textbf{Thamar Solorio$^\ddagger$} \\
$^\ddagger$University of Houston, Houston, TX \\
$^\dagger$Salesforce Research, Palo Alto, CA \\
$^\ddagger$\texttt{\{gaguilaralas, tsolorio\}@uh.edu} \\
$^\dagger$\texttt{\{bmccann, tniu, nazneen.rajani, nkeskar\}@salesforce.com}
}

\begin{document}
\maketitle

\renewcommand{\thefootnote}{\fnsymbol{footnote}}
\footnotetext[1]{Work performed as summer intern at Salesforce.}
\footnotetext[7]{Work performed as manager while at Salesforce.} 
\renewcommand*{\thefootnote}{\arabic{footnote}}

\begin{abstract}
Byte-pair encoding (BPE) is a ubiquitous algorithm in the subword tokenization process of language models as it provides multiple benefits.
However, this process is solely based on pre-training data statistics, 
making it hard for the tokenizer to handle infrequent spellings. 
On the other hand, though robust to misspellings, pure character-level models often lead to unreasonably long sequences and make it harder for the model to learn meaningful words.
To alleviate these challenges, we propose a character-based subword module (char2subword)\footnote{The code is available at \url{https://github.com/salesforce/char2subword}} that learns the subword embedding table 
in pre-trained models like BERT.
Our char2subword module builds representations from characters out of the subword vocabulary, and it can be used as a drop-in replacement of the subword embedding table. 
The module is robust to character-level alterations such as misspellings, word inflection, casing, and punctuation. 
We integrate it further with BERT through pre-training while keeping BERT transformer parameters fixed--and thus, providing a practical method. 
Finally, we show that incorporating our module to mBERT significantly improves the performance on the social media linguistic code-switching evaluation (LinCE) benchmark. 
\end{abstract}
\section{Introduction}



\begin{figure}[t!]
\centering
\includegraphics[width=0.95\linewidth]{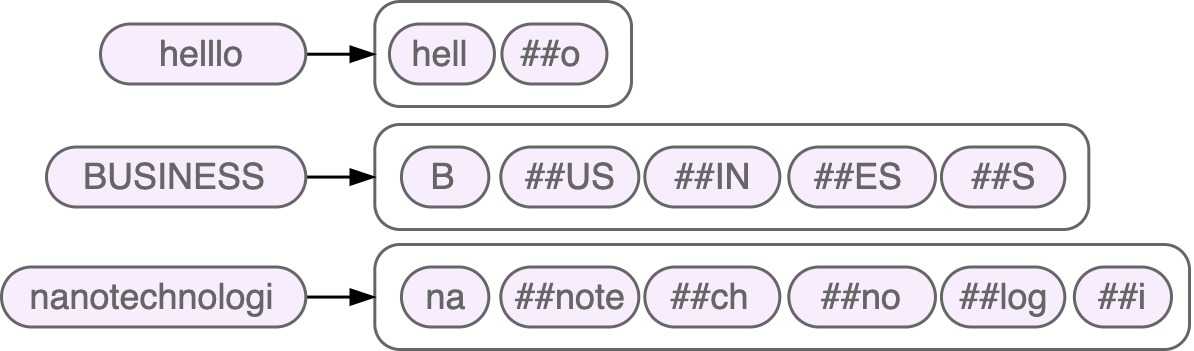}
\caption{
\label{fig:bpe_example}
Examples of subword tokenization from OOV words. 
The word \textit{helllo} changes its meaning (e.g., \textit{hell}), \textit{BUSINESS} is split almost to characters, and \textit{nanotechnologi} do not resemble any of its morphemes.
}
\end{figure}

Byte-pair encodings (BPE) is a ubiquitous algorithm in the tokenization process among transformer-based language models such as 
BERT \citep{devlin-etal-2019-bert}, 
GPT-2 \citep{radford2019language}, 
RoBERTa \citep{liu-etal-2019-roberta}, 
and CTRL \citep{keskar-etal-2019-ctrl}.
This method addresses the open vocabulary problem by segmenting unseen or rare words into smaller subword units while keeping a reasonable vocabulary size \citep{huck-etal-2017-target,kudo-2018-subword,wang2018multilingual}.
However, BPE and its variants are sensitive to small perturbations in the text, potentially 
distorting the sentences' meaning \citep{jones-etal-2020-robust} (see Figure \ref{fig:bpe_example}).
Moreover, this tokenization process is rigid to changes such as adding more subwords to the vocabulary or correcting the segmentation splits.
That is because the tokenization relies on the original corpus where the vocabulary was generated (e.g., Wikipedia), resulting in a fixed set of subword pieces tied to an embedding lookup table \citep{bostrom-durrett-2020-byte}.
Although these aspects are not a problem with clean and properly formatted text, that is not the case when the text presents substantial noise (e.g., Wikipedia vs. social media). 
Noisy text can result in extensive subword pieces per word (see Figure \ref{fig:bpe_example}), preventing the models from capturing the meaning effectively and adapting to such domains.
This is particularly prominent on social media text \citep{baldwin-etal-2015-shared,eisenstein-2013-phonological,eisenstein-2013-bad}, where the noise permeates even across languages and in code-switching scenarios \citep{singh-etal-2018-twitter,aguilar-etal-2018-named,molina-etal-2016-overview,das-codemixing-contest-icon2016}.

This paper proposes a character-to-subword (char2subword) module trained to handle rare or unseen spellings robustly while being less restrictive to a particular tokenization method.
Our method works as a drop-in alternative to the embedding table in pre-trained language models like mBERT. 
It improves performance and reduces the number of embedding parameters by 45\% without sacrificing inference speed.
We train our module to approximate the embedding table using characters from the original vocabulary words and subwords. 
This procedure leverages transfer learning from the pre-trained embedding table rather than starting from scratch---thus, saving precious computational time and resources.
Besides, the subword vocabulary provides enough character-level patterns to learn from already-segmented tokens.
We integrate our module with mBERT's transformer layers even further by continuing to train with the pretraining data and the MLM objective.
Once our char2subword is adapted to the pre-trained language model, we evaluate the overall model performance by fine-tuning it on downstream tasks.
We show our method's effectiveness by outperforming mBERT on the social media linguistic code-switching (LinCE) benchmark \citep{aguilar-etal-2020-lince}, where the fine-tuning domain deviates substantially from the pre-training domain.
The results show that the char2subword module can also capture intra-word code-switching.
At the sentence level, the model can relate words from the same language to support language prediction.

We highlight our main contributions as follows:
\begin{enumerate}[topsep=0pt,itemsep=-1ex,partopsep=1ex,parsep=1ex]
\item 
We introduce char2subword, a new parameter-efficient and open-vocabulary module that extends the domain-constrained and fixed vocabulary in mBERT (or any pre-trained model relying on subwords) while preserving the semantics of the multilingual embedding space. 
\item 
We show the character compositionality capabilities of our module by handling noise robustly at the character level while being language-independent and flexible to different tokenization.
\item 
We analyze the advantages of our model on downstream tasks and demonstrate its practical use and adaptability to other domains despite of vocabulary changes.
\end{enumerate}{}


\section{Related Work}



\paragraph{Word representations}
Most of the initial ground-breaking advances in NLP relied on word embedding representations from methods like word2vec \citep{mikolov2013distributed} and GloVe \citep{pennington-etal-2014-glove}. 
They showed the capability of arranging words in a continuous high-dimensional space encoding semantic relationships and meaning \citep{DBLP:journals/corr/GoldbergL14}. 
However, rare words are weakly represented in such space, and OOV words are not representable.
To alleviate that, researchers proposed word representations using recursive neural networks guided by morphology \citep{luong-etal-2013-better}, as well as morpheme embeddings as a prior distribution over probabilistic word embeddings \citep{bhatia-etal-2016-morphological}.
Regardless, the challenges persist in noisy text, where users do not follow the canonical word forms \citep{eisenstein-2013-bad}. 
Such problems are aggravated in social media due to the inherently multilingual environment. 
More words per language are required, while the spelling noise is persistent across languages. 


\paragraph{Character representations}
While character-level systems proved strong for text classification \citep{conneau-etal-2017-deep}, they were not as successful on multilingual tasks like neural machine translation (NMT) initially \citep{neubig2013substring,chung-etal-2016-character}.
Even when the performance was satisfactory, such systems had to process long sequences of characters resulting in a very slow process
\citep{costa-jussa-fonollosa-2016-character,DBLP:journals/corr/LingTDB15}.
Additionally, languages have different writing systems and specific properties encoded at the character level. 
While some of those properties may be captured effectively on morphologically rich languages (e.g., Czech and Arabic), properties from other languages are not more impactful than using words (e.g., English) \citep{cherry-etal-2018-revisiting}.
These challenges are also applicable to our case since we conduct our study on multilingual data with typologically different languages.


\paragraph{Hybrid representations}
Using words or characters has shown advantages and disadvantages on both ends. 
Researchers tried to get the best of both worlds by combining characters and words in a hybrid architecture \citep{luong-manning-2016-achieving} where the default was based on static word embeddings that backed off to characters if the word was unknown. 
Parallel efforts focused on character-aware neural language models \citep{Kim2016CharacterAwareNL} where the meaning is contextually enriched by highway networks \citep{DBLP:journals/corr/SrivastavaGS15}, as well as character-based LSTM language models that build intermediate word representations from character-level LSTMs \citep{ling-etal-2015-finding}. 
Most successful contextualized word embeddings built out of characters are the language models ELMo \citep{peters-etal-2018-deep} and Flair \citep{akbik-etal-2018-contextual}.
Building models from characters can easily adapt to social media domains \citep{akbik-etal-2019-pooled}, including code-switching data \citep{aguilar-solorio-2020-english}.

\paragraph{Subword models}
\citet{sennrich-etal-2016-neural} proposed subword tokenization using the byte-pair encoding (BPE) algorithm to balance the use of characters and words.
BPE automatically chooses a vocabulary of subwords given the desired vocabulary size.
This procedure recursively builds subwords upon characters using the word frequencies \citep{sennrich-etal-2016-neural}. 
Another greedy variation of BPE can select the longest prefix to segment words \citep{DBLP:journals/corr/WuSCLNMKCGMKSJL16}.
Alternatively to the greedy versions, the segmentation can happen in a stochastic way; 
drawing segmentation candidates at different points of a word can improve generalization \citep{kudo-2018-subword}.
The WordPiece variation of BPE is used in NMT 
and language models such as multilingual BERT \citep{devlin-etal-2019-bert}.
Regardless of the variant, these methods handle the out-of-vocabulary problem by breaking down unseen or rare words into pieces that are in the vocabulary. 
The problem is that BPE can generate subword pieces that are not linguistically plausible.
The BPE tokenization is not ideal for social media domains because its rules do not necessarily apply across domains, particularly the ones with substantial noise and spelling differences \citep{bostrom-durrett-2020-byte}.

\paragraph{Compositional models}
The idea of composing OOV vectors has been explored before \citep{ling-etal-2015-finding, plank-etal-2016-multilingual}. 
However, learning such vectors requires a large corpus and long computing time (i.e., processing characters). 
\citet{pinter-etal-2017-mimicking} proposed learning OOV words from a pre-trained word embedding dictionary. 
They treat every word from the dictionary as a sequence of characters and output a single vector that mimicks the associated word embedding in the dictionary. 
\citet{schick-schutze-2019-attentive} improved this method by introducing \textit{attentive mimicking} to account for context, besides the surface form of the word. 




%
%
%
%
%

\section{Method}
Given a word $w$, a subword model produces a sequence of subword pieces $\vs = (s_0, s_1, \dots, s_n)$, such that the concatenation of all the segments from $\vs$ fully reconstructs the word $w$. 
Regardless of whether a subword piece represents a character in a word or not, all the pieces are treated as semantic units within a sentence.\footnote{Previous studies \citep{clark-etal-2019-bert,rogers-etal-2020-primer} showed that BERT learns syntax and parsing within its self-attention probabilities. That is evidence that subwords need to preserve semantics when fed into such layers. This suggests that subword pieces broken down to the character level can prevent the model from exploiting linguistic properties.} 
Such pieces come from a rule-based system that does not take into account semantics or morphology during the tokenization.
Thus, the subword tokenization has a significant impact on the semantic abstraction from upper layers in pre-trained models like mBERT.

To alleviate such problems, we build word representations out of characters.
The char2subword module allows flexible tokenization patterns, where the model can split by spaces, use the original tokenization method, or employ a different tokenization process as defined by the user.
There are two main phases in our proposed method: approximating 
subword embeddings with the char2subword module (i.e., ideally replicating the embedding space $\mE$) (Section \ref{ch9:sec:ch2sw_mimic}), and contextually integrating the char2subword module into the pre-trained model (Section \ref{ch9:sec:ch2sw_pretrain}).

\subsection{Approximating the subword embedding} 
\label{ch9:sec:ch2sw_mimic}


Consider a subword $s_i$ from the vocabulary $\mathcal{V}$ and a subword embedding matrix $\mE \in \sR^{|\mathcal{V}|\times d}$. 
We learn a parameterized function $f_\theta: \sR^{|c| \times 1} \rightarrow \sR^d$ that maps the sequence of characters $\vc_i = (c_{i1}, c_{i2}, \dots, c_{i|s_i|})$ from the subword $s_i$ to its corresponding embedding vector $\ve_i \in \mE$:
\begin{align*}
    \hat{\ve_i} = f_\theta(\vc_i) ~~~s.t.~~ \hat{\ve}_i \approx \ve_i
\end{align*}

To accomplish this, we design an objective function that fulfills our four desiderata; we want the embeddings to: 
(i) preserve their angular distances,
(ii) be similar in $\normltwo$ norm to prevent magnitude disruptions in upper layers of mBERT, 
(iii) have similar neighbors in cosine-distance space, and 
(iv) ultimately map to the same tokens in embedding space. We thus optimize $f_\theta$ by minimizing the overall objective function $\mathcal{L}(\cdot)$:
\begin{gather+}[0.93]
\mathcal{L}(\vc_i, \ve_i, \vy_i, f_{\theta}) =
    \mathcal{L}_{cos}(\ve_i, f_{\theta}(c_i)) + \normltwo(\ve_i, f_{\theta}(c_i)) \\
    ~~~~~~~~~~~~~~~~~~~~~~~~~~~
    + \mathcal{L}_{nbr}(\ve_i, f_{\theta}(c_i)) + 
    \mathcal{L}_{ce}(\vy_i, f_{\theta}(c_i))
\end{gather+}

The four objectives of the loss function correspond to the four aforementioned desired properties. 
The first objective, $\mathcal{L}_{cos}(\cdot)$, is the cosine distance between the target and the predicted embedding vectors $\ve_i$ and $\hat{\ve}_i$. 
By using an angular distance function, we encourage the model to replicate the semantic relationships and vector arrangements in the original embedding space of $\mE$:
\begin{align*}
    \mathcal{L}_{cos}(\ve_i, \hat{\ve}_i) = ~1 - \frac{\ve_i \cdot \hat{\ve}_i}{||\ve_i|| ~||\hat{\ve}_i||}
\end{align*}

The second objective is the $\normltwo$ norm or euclidean distance between the vectors $\ve_i$ and $\hat{\ve}_i$.
The previous objectives do not regulate the magnitude of the predicted vector $\hat{\ve}_i$, allowing that to be a degree of freedom for $f_\theta$.
By using the $\normltwo$ norm, we penalize the model for generating a vector $\hat{\ve}_i$ with a different magnitude than $\ve_i$. 
Regulating the magnitude is important to approximate the vector arrangements in the embedding space. 
We hypothesize that slightly different properties in the embedding $\mE$ can magnify differences at the upper layers of mBERT.

The third objective, $\mathcal{L}_{nbr}(\cdot)$, is the mean squared error (MSE) of cosine distances
generated between the $k$-th closest neighbors to $\ve_i$ versus the distances of the same neighbors with respect to $\hat{\ve}_i$:
\begin{gather+}[0.93]
    (\vn_1, \dots \vn_k) =~ topk(\ve_i, \mE) ~~~~~~~~~~~~~~~~~~~~~~~~~~~~~~~~~~~~~\\
    \mathcal{L}_{nbr}(\ve_i, \hat{\ve}_i) =~ \frac{1}{k}\sum_{j=1}^k{( dis(\ve_i, \vn_j) - dis(\hat{\ve}_i, \vn_j)  )^2}
\end{gather+}
where $topk(\cdot, \cdot)$ retrieves the $k$-th closest neighbors according to the cosine distances among all the subword vectors in $\mE$.
The core idea of this objective is to force distances between $\hat{\ve}_i$ and the neighbors $\vn_*$ to be as similar as possible to the distances between the same neighbors and $\ve_i$.

The final objective is the cross-entropy loss $\mathcal{L}_{ce}(\cdot)$.
We use $\mE$ as fixed parameters to project linearly from the embedding to the vocabulary.
This loss term forces the model to learn accurate embedding representations such that they map to the original subwords from the vocabulary $\mathcal{V}$:
\begin{align*}
\mathcal{L}_{ce}(\vy_i, \hat{\ve}_i) &= -\sum^{|\mathcal{V}|}_j \vy_{ij} \log \hat{\vy}_{ij} \\
s.t. ~~~\hat{\vy}_i &= \softmax(\hat{\ve}_i \cdot \mE^\top)
\end{align*}

\begin{figure}[t!]
\centering
\includegraphics[width=\linewidth]{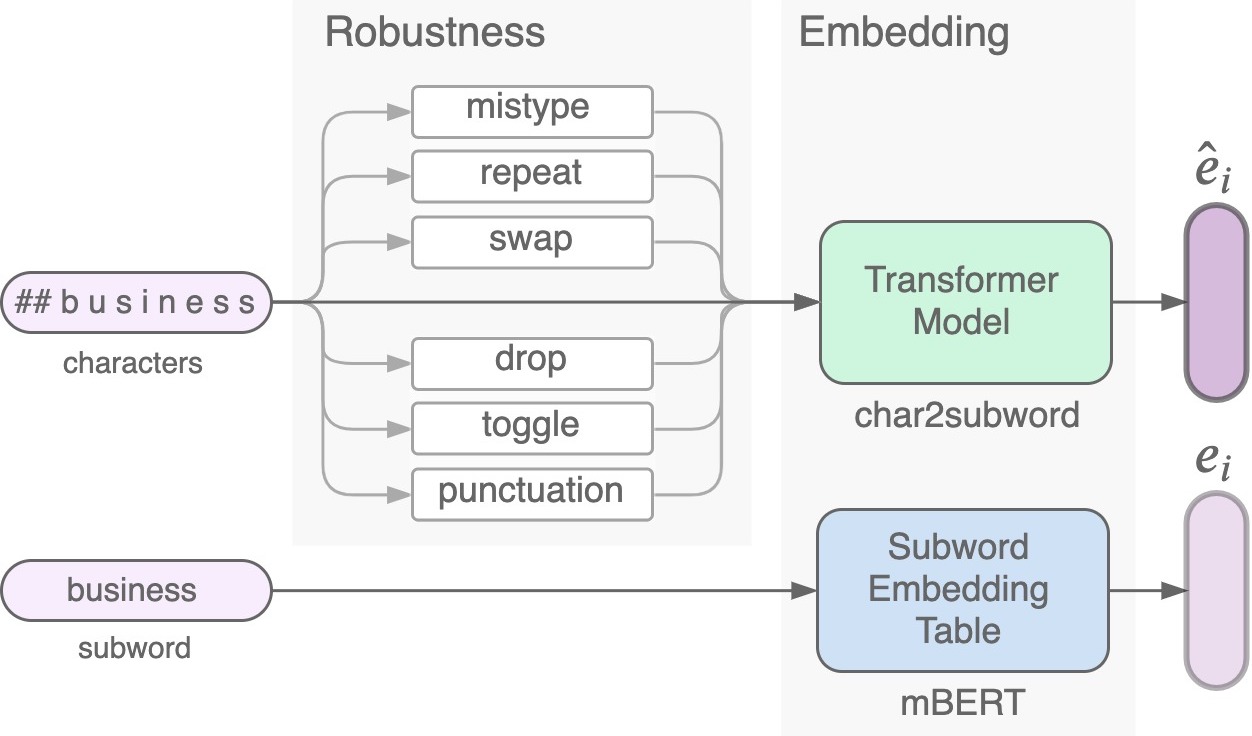}
\caption{
The char2subword module approximates the mBERT subword embedding table.
We incorporate noise in every word with single-character operations. 
}
\label{ch9:fig:char2subword}
\end{figure}
\begin{table}[t!]
\small
\centering
\resizebox{\linewidth}{!}{
\begin{tabular}{lp{2.2in}}
    \toprule
    \bf Operation & \bf Description \\
    \midrule
    \texttt{mistype} & replace a random character of a subword by randomly choosing from its nearby keys according to a keyboard layout \\
    \midrule
    \texttt{repeat} & repeat a random character of a subword \\
    \midrule
    \texttt{swap} & randomly choose a character and swap it with the next character in a subword \\
    \midrule
    \texttt{drop} & randomly drop a character of a subword \\
    \midrule
    \texttt{toggle} & toggle the case of a randomly chosen character from a given subword \\
    \midrule
    \texttt{punct} & randomly add a punctuation mark commonly used within words (e.g., dashes, periods) \\
    \bottomrule
\end{tabular}
}
\caption{
Single-character operations to incorporate noise in the approximation stage. 
The operations are applied to every word in the vocabulary that exceeds the four characters. 
}
\label{ch9:tab:noise_operations}
\end{table}

\paragraph{Char2subword module} 
We model $f_\theta$ using the transformer architecture \citep{vaswani2017attention}.
The module processes a sample as a sequence of characters $\vc_i = (c_{i1}, c_{i2}, \dots, c_{iM})$ 
of a subword $s_i$ of length $M$.\footnote{
To distinguish between words and subwords, we prepend `\#\#' to the sequence $\vc_i$ in the case of full words.
}
We represent the sequence $\vc_i$ as the sum between the character embeddings and sinusoidal positional encodings.
We pass the resulting sequence of character vectors $\mX_0$ to a stack of $l$ attention layers, each with $k$ attention heads. On top of the $l$ attention layers, we add a linear layer $\mW_e \in \sR^{d' \times d}$ followed by max-pooling and a layer normalization for the final output $\hat{\ve}_i$ (see full definition in Appendix \ref{app:char2subword_definition}).

\paragraph{Character-level robustness}
The flexibility of the char2subword module makes it easier to teach the model text invariance because the inputs are now processed at the character-level.
We augment the subword vocabulary $\mathcal{V}$ by introducing natural single-character misspellings during training. 
We apply one operation at a time and only to subwords that exceed four characters to reduce the chance of ambiguity between valid subwords.
The operations are described in Table \ref{ch9:tab:noise_operations} and the high-level view of the approximation appears in Figure \ref{ch9:fig:char2subword}.\footnote{
For the \texttt{mistype} operation, we use over 100 keyboard layouts to cope with the languages in mBERT.
}

\subsection{Pre-training with the char2subword}
\label{ch9:sec:ch2sw_pretrain}

The previous techniques leverage the pre-trained knowledge in the embedding matrix $\mE$. 
However, the char2subword module may not be integrated with the pre-trained mBERT's upper layers since it has only seen individual subwords without context.
To alleviate that, we pre-train the char2subword module along with mBERT \citep{gururangan-etal-2020-dont}. 
We do not update parameters in upper layers of mBERT since the goal is to provide the char2subword module as a drop-in alternative for $\mE$ on the publicly available pre-trained models.\footnote{
While the study focuses on mBERT, this method can be applied to other pre-trained models like RoBERTa or XLM-R.
}

\begin{figure}[t!]
\centering
\includegraphics[width=\linewidth]{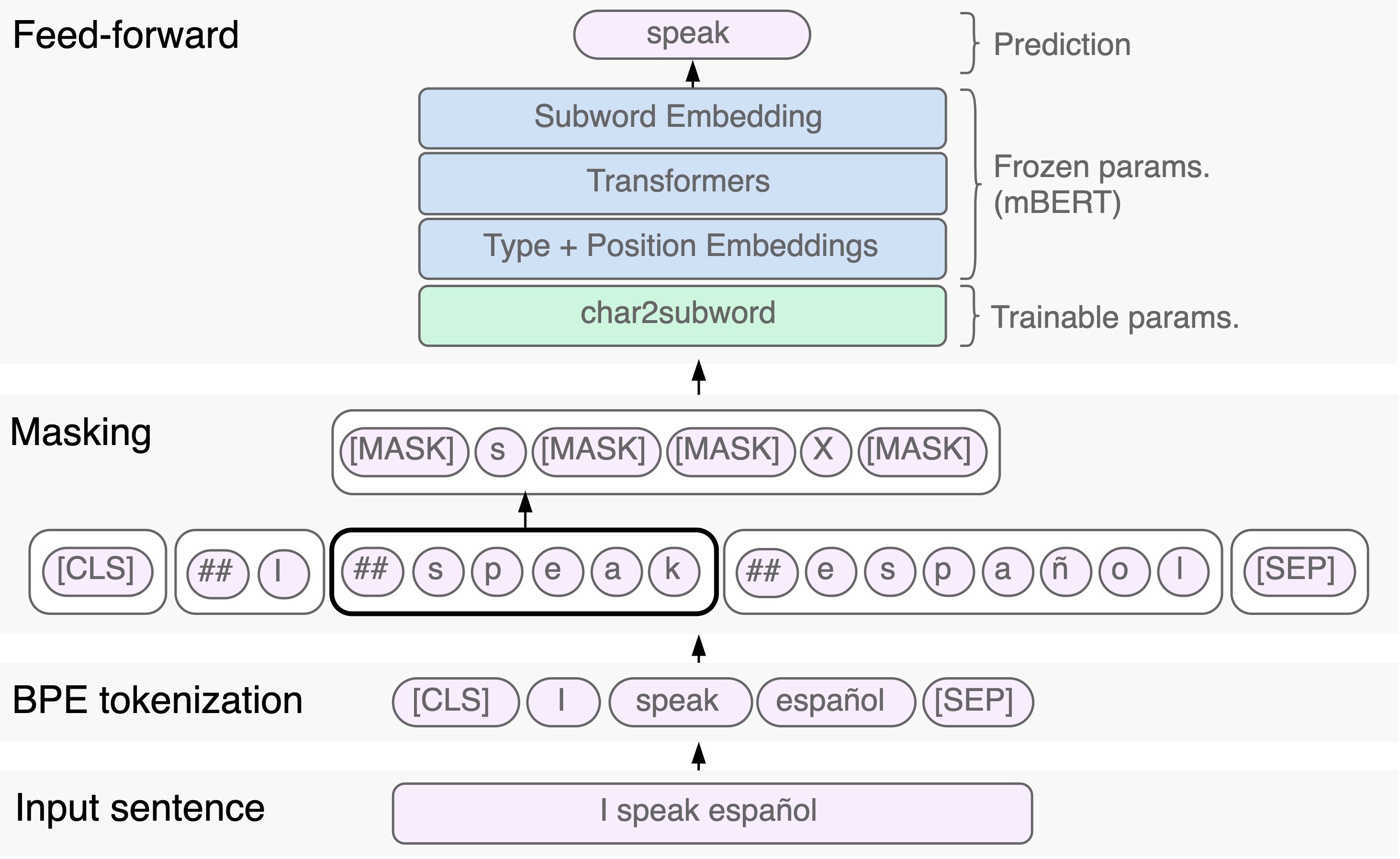}
\caption{
An example of an input and output of the pre-training setting with a masked language modeling (MLM) objective at the character level.
}
\label{ch9:fig:pretraining}
\end{figure}

Following \citet{liu-etal-2019-roberta}, we use a dynamic masked language modeling (MLM) objective (see Figure \ref{ch9:fig:pretraining}).
We randomly choose 15\% of the subword tokens and mask them at the character level. 
We replace 80\% of the characters with \texttt{[MASK]}, 10\% with randomly chosen characters and the remaining 10\% is left unchanged.
We feed characters to the char2subword module and make predictions from the subword vocabulary $\mathcal{V}$.\footnote{
We project the internal representations per word onto the vocabulary space using $\mE$ (without updating its parameters).
}
We pre-train the char2subword model with 1M sequences of 512 subword tokens from Wikipedia (200K sequences for each English, Spanish, Hindi, Nepali, and Arabic text).
Using gradient accumulation, we update parameters with an effective batch size of 2,000 samples.
Note that the model does not require extensive pre-training since 1) the upper-layer parameters are initialized from the pre-trained mBERT checkpoint and kept fixed during training, and 2) the char2subword module is initialized from the embedding approximation phase.
Thus, pre-training the model for a few epochs is sufficient.

\subsection{Fine-tuning}
\label{ch9:sec:ch2sw_finetune}

Once the char2subword module has been optimized, we evaluate the pre-trained model with the char2subword module on downstream NLP tasks.
Specifically, we experiment with two scenarios: the full and the hybrid modes. 

\paragraph{Full mode} 
This mode completely replaces the subword embedding table in mBERT (i.e., the set of parameters and vectors) with the char2subword module.
The idea of this setting is to evaluate how well approximated was the embedding space originally in $\mE$.
Intuitively, if the char2subword replicates the embedding space in $\mE$ perfectly, then the overall model should behave about the same as the original mBERT model.
Nevertheless, this setting does not tokenize further a word; hence, the input sequence tends to be shorter and more meaning-preserving (i.e., too many subword pieces for a single word can degrade its meaning).


\paragraph{Hybrid mode} 
Unlike the full mode, this mode does not replace the subword embedding table. 
Instead, it uses the subword embedding vectors by default for full words (i.e., not subword pieces). The model backs off to character-based embeddings from the char2subword module when a word as a whole does not appear in the vocabulary.
This method focuses specifically on subwords rather than full words, effectively preventing words from being broken down into pieces.


\section{Experiments}

\paragraph{Embedding approximation}
The goal of the approximation experiments is to replicate the original subword embedding table while ensuring robustness at the character level.
We experiment with the objective functions described in Section \ref{ch9:sec:ch2sw_mimic}.
We use the average precision to determine the best method (we also provide the accuracy for reference).\footnote{
Using accuracy to determine the best method can mislead the interpretation of the model's capabilities. 
Accuracy is not ideal in this scenario since the goal is to approximate an embedding space rather than merely predicting vocabulary subwords given their characters.
}
The experiments 1.1-1.4 show the results of each objective separately (see Table \ref{tab:simulation_dev_results}). 
Notably, the cross-entropy objective is the most relevant to ensure high precision (58\% vs. 28.5\% of the cosine objective).
Combining all the objectives gives an average precision of 60\% (experiment 1.9). 
Although experiment 1.6 and 1.9 perform very close (59.9\% vs. 60\%), the latter still preserves more neighbors along the top $k$ expected neighbors.

\begin{table}[t!]
\centering
\resizebox{\linewidth}{!}{
\setlength{\tabcolsep}{3pt}
\begin{tabular}{lcccccccccc}
\toprule
\multicolumn{1}{l}{\textbf{Exp.}} 
    & \textbf{$\mathcal{L}_{ce}$}
    & \textbf{$\mathcal{L}_{cos}$} 
    & \textbf{$\normltwo$}
    & \textbf{$\mathcal{L}_{nbr}$} 
    & \textbf{Acc.} 
    & \textbf{Prec@1} 
    & \textbf{Prec@15} 
    & \textbf{Avg Prec} \\
\midrule
1.1  & \cmark & & & & 99                   & 99.6  & 43.9  & 58.1   \\
1.2  & & \cmark & & & 62                   & 41.8  & 24.2  & 28.5   \\
1.3  & & & \cmark & & 45                   & 18.2  & 12.2  & 13.5   \\
1.4  & & & & \cmark & 43                   & 25.5  & 17.1  & 19.6   \\
\midrule
1.5   & \cmark & \cmark & & & 96           & 96.1  & 41.2  & 55.1   \\
1.6   & \cmark & & \cmark & & 95           & 99.1  & 46.6  & 59.9   \\
1.7   & \cmark & \cmark & \cmark & & 95    & 98.6  & 46.7  & 59.8   \\
1.8   & \cmark & & & \cmark & 98           & 97.4  & 42.6  & 56.5   \\
\midrule
1.9 & \cmark  & \cmark  & \cmark  & \cmark  & 95   & 98.3  & 47.1  & \textbf{60.0} \\
\bottomrule
\end{tabular}
}
\caption{\label{tab:simulation_dev_results} 
    The results of approximating the subword embedding table from mBERT using different objectives (\cmark).
    %
    %
    The accuracy denotes the capability of the model to predict a subword out of its characters.
    Precision @ $k$ measures the overlap between the $k$ ground-truth neighbors for a vector $\ve_i$ (that represents subword $s_i$) and the $k$ neighbors of the predicted vector $\hat{\ve}_i$.
}
\end{table}

\begin{figure}
\centering
\includegraphics[width=\linewidth]{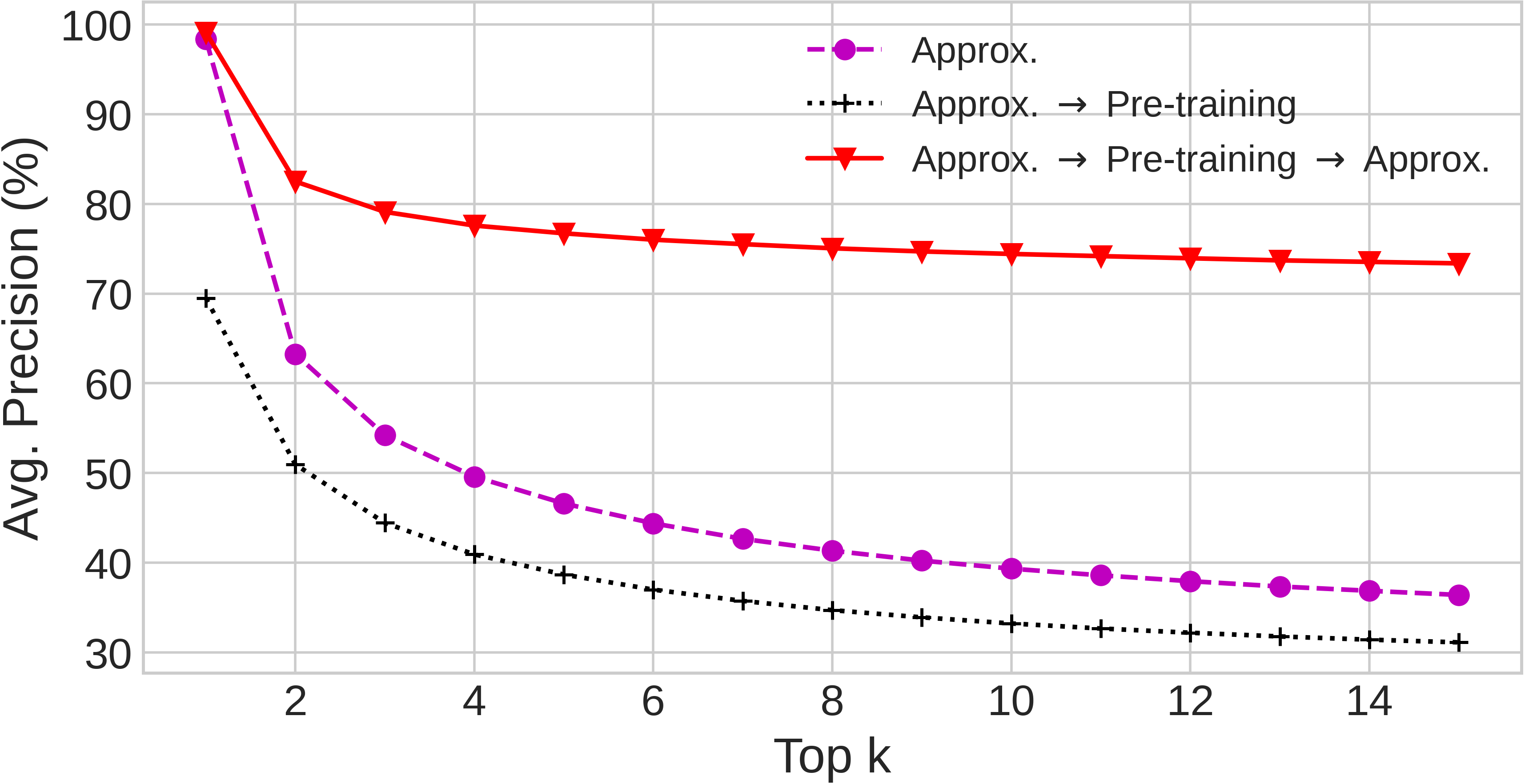}
\caption{
\label{ch9:fig:char2subword_precision:b}
The precision up to 15 neighbors combining the approx. and pre-training phases in different ways.
}
\end{figure}

After optimizing a char2subword module (experiment 1.9), we contextualize it in the pre-training phase (Section \ref{ch9:sec:ch2sw_pretrain}).
The results show that the precision at $k$ drops substantially (Figure \ref{ch9:fig:char2subword_precision:b}, ``Approx. $\rightarrow$ Pre-training'').
However, when restarting approximation after pre-training, the model performs far better than the initial approximated version reaching an average precision of 82.4\% (Figure \ref{ch9:fig:char2subword_precision:b}, ``Approx. $\rightarrow$ Pre-training $\rightarrow$ Approx.'').
This improvement shows the need for contextualization for the original char2subword module. 
Contextualization by itself does not guarantee that the module will resemble the same embedding space as in $\mE$ (i.e., nothing that forces the module to optimize for that). 
However, it aligns better the semantics of the space facilitating the approximation.

Character-level robustness is another essential aspect when optimizing the char2subword.
We add single-character edits to the training phase described in Section \ref{ch9:sec:ch2sw_mimic}.
Table \ref{tab:simulation_neighbors} shows the neighbors of the word \textit{business} and its variations.
When fed \textit{business} without noise, the char2subword modules (with and without noise) retrieve semantically-related neighbors.
However, when the word is capitalized, the neighbors are not related to the word \textit{business} for the char2subword without noise.
Also, the subword tokenization for \textit{BUSINESS} becomes \textit{B-US-INE-SS}, which distorts the meaning of the original word.
Regardless, the char2subword with noise is resilient to the capitalization pattern and capable of maintaining the meaning.\footnote{
The char2subword module never sees a subword from the vocabulary with more than a single character edit (i.e., we defined the robustness procedure this way). 
That means that the word \textit{BUSINESS} never appeared in training for the model.
}

\begin{table*}[t!]
\centering
\small
\resizebox{\linewidth}{!}{
\begin{tabular}{@{}lll@{}}
\toprule
\textbf{Input}
& \textbf{Model}
& \textbf{Neighbors} \\ 
\midrule
\multirow{3}{*}{business} 
    & mBERT                & (business, 1.0), (Business, 0.61), (businesses, 0.47), (businesses, 0.47), (bisnis, 0.46) \\
    & Char2subword         & (business, 0.82), (Business, 0.50), (businesses, 0.43), (\foreignlanguage{russian}{бизнес }, 0.40), (bisnis, 0.38)    \\
    & Char2subowrd + noise & (business, 0.80), (Business, 0.61), (businesses, 0.53), (\foreignlanguage{russian}{бизнес}, 0.43), (negocios, 0.39)  \\
\midrule
\multirow{2}{*}{bsusinessses} 
    & Char2subword         & (businesses, 0.42), (companies, 0.33), (opportunities, 0.32), (industries, 0.31) 
    \\
    & Char2subword + noise & (businesses, 0.79), (companies, 0.53), (shops, 0.52), (corporations, 0.50), (employees, 0.49)    \\
\midrule
\multirow{2}{*}{BUSINESS} 
    & Char2subword         & (ASEAN, 0.25), (RSS, 0.24), (FCC, 0.24), (WEB, 0.2403), (Austrália, 0.2360)                      \\
    & Char2subword + noise & (Business, 0.53), (business, 0.32), (Marketing, 0.31), (Corporate, 0.31), (Communications, 0.30) \\
\bottomrule
\end{tabular}
}
\caption{\label{tab:simulation_neighbors}
Neighbors from the mBERT subword embedding table using different embedding vectors to represent the word \textit{business} and its modifications (e.g., $topk(\ve, \mE)$) .
For the mBERT OOV words \textit{bsusinessses} and \textit{BUSINESS}, the tokenizer breaks the words as \textit{b-sus-iness-ses} and \textit{B-US-INE-SS}, respectively.
}
\end{table*}

\paragraph{Fine-tuning experiments}
Once the module is adapted to mBERT, we benchmark the model in the full and hybrid modes (see Section \ref{ch9:sec:ch2sw_finetune}) using the LinCE benchmark \citep{aguilar-etal-2020-lince}.
Particularly, we focus on language identification (LID), part-of-speech (POS) tagging, named entity recognition (NER), and sentiment analysis (SA).
Table \ref{ch9:tab:lince_dev_results} shows the results of the experiments using the full and hybrid modes.
Also, we include ELMo's test scores\footnote{\url{ritual.uh.edu/lince/leaderboard}} as baseline since ELMo composes its representations from characters.
For each proposed model, we use the approximated and pre-trained (i.e., ``Approx. $\rightarrow$ Pre-training $\rightarrow$ Approx.'') versions of the char2subword module.
The language identification results are not a strong indicator of improvement since the scores are all very close.\footnote{
The average score for LID across language pairs is 95.71\% for mBERT (baseline) and 95.80\% for char2subword module (hybrid, pre-trained).
} 
Nevertheless, it is important to note that the model, regardless of the version, can perform on par with the mBERT baseline.
This suggests that the char2subword representations are compatible with the rest of the mBERT model (i.e., mBERT transformer layers).

\begin{table*}[t!]
\centering
\small
\resizebox{\textwidth}{!}{
\begin{tabular}{llccccccccccc}
\toprule
& & & \multicolumn{4}{c}{\textbf{LID (W. F\sub{1})}}
& \multicolumn{2}{c}{\textbf{POS (Acc.)}}
& \multicolumn{3}{c}{\textbf{NER (F\sub{1})}} 
& \multicolumn{1}{c}{\textbf{SA (W Acc.)}} 
\\
\cmidrule(lr){4-7}
\cmidrule(lr){8-9}
\cmidrule(lr){10-12}
\cmidrule(lr){13-13}
\textbf{Method} 
        & \textbf{Adaptation} & \textbf{Avg}
        & es-en & hi-en & ne-en & msa-arz 
        & es-en & hi-en  
        & es-en & hi-en & msa-arz  
        & es-en
\\  
\midrule
\multicolumn{2}{l}{\textit{Validation set results}}
\\
~~~~mBERT
        & N/A 
        & 83.86
        & 98.23  & 96.37  & \textbf{96.67}  & 91.55 
        & \textbf{97.29}  & 87.86   
        & 62.66  & 72.94  & 78.93  
        & 56.10
\\
\cmidrule(lr){3-13}
~~~~Full    & Approx.     
        & 83.59 
        & 98.16 & 95.79 & 96.45	& 91.63	
        & 96.93	& 89.04	
        & 62.02	& 70.79	& 79.13 
        & 55.98 %
\\
~~~~Full    & Pre-trained   
        & 83.89   
        & 98.20 & 96.97 & 96.47	& 91.48	
        & 96.91	& 89.38	
        & 61.23	& 71.98	& 79.42 
        & 56.82 %
\\
~~~~Hybrid  
        & Approx.\makebox[0pt][l]{$^{\ast}$}    
        & 84.33            
        & \textbf{98.24} & \textbf{96.98}	& 96.50	& 91.48	
        & 97.16	& 88.95	
        & \textbf{64.26}& 72.68	& 80.10 
        & 56.98 %
\\
~~~~Hybrid  
        & Pre-trained\makebox[0pt][l]{$^{\ast}$} 
        & \textbf{84.60}  
        & 98.18	& 96.75	& 96.37	& \textbf{91.64}
        & 97.03	& \textbf{89.64}
        & 63.32	& \textbf{74.91} & \textbf{80.45} 
        & \textbf{57.71} %
\\
\midrule
\multicolumn{2}{l}{\textit{Test set results}}
\\
~~~~ELMo
        & N/A
        & 79.52 
        & 97.93	& 95.43	& 95.90	& 86.53
        & 96.34	& 86.71
        & 52.58	& 68.79	& 56.68
        & 52.88
\\
~~~~mBERT
        & N/A 
        & 82.23 
        & \textbf{98.36}  & 94.24  & \textbf{96.32}  & \textbf{91.55}
        & \textbf{97.07}  & 86.30 
        & 64.05  & 72.57  & 65.39  
        & 56.43
\\
\cmidrule(lr){3-13}
~~~~Hybrid  
        & Pre-trained\makebox[0pt][l]{$^{\ast}$} 
        & \textbf{83.03} 
        & 98.33  & \textbf{96.23}  & 96.19  & 91.19 
        & 96.88  & \textbf{88.23}  
        & \textbf{64.65}  & \textbf{73.38}  & \textbf{66.13}  
        & \textbf{59.07}
\\
\midrule
\multicolumn{12}{l}{{\small * Statistically significant with respect to the mBERT baseline, with $p$-value $< 0.01$ in student's t-test \citep{dror-etal-2018-hitchhikers}.}} 
\\
\bottomrule
\end{tabular}}
\caption{
Results on the LinCE benchmark. 
Full refers to the full mode where the model only uses the char2subword to embed the input.
Hybrid means that the model uses the subword embedding table by default and backs off to the char2subword module for OOV words, instead of splitting them.
For this table, pre-trained means that the model was approximated after the pre-training phase (i.e., ``Approx. $\rightarrow$ Pre-training $\rightarrow$ Approx.'').
The languages involved are English (en), Spanish (es), Hindi (hi), Nepali (ne), Modern Standard Arabic (msa), and Egyptian Arabic (arz).
The best results on each language pair are in bold, and the test scores are in italics.
}
\label{ch9:tab:lince_dev_results}
\end{table*}

For the POS and NER tasks, we see improvements compared to mBERT. 
The hybrid pre-trained experiment for Hindi-English is significantly better than the baseline for both POS (89.64\% vs. 87.86\%) and NER (74.91\% vs. 72.94\%).
One of the reasons for this performance boost is due to the noise that splitting transliterated Hindi (i.e., Romanized Hindi) generates for the baseline. 
On the contrary, the char2subword compresses the transliterated words into a single vector, reducing the noise in the model.
The NER results for Spanish-English (es-en) and Modern Standard Arabic-Egyptian Arabic (msa-arz) also exceed the baseline (64.26\% vs. 62.66\%). 
Although there is no transliteration in these language pairs, there is still much noise coming from social media user-generated language.
Also, pre-training the char2subword on Spanish and Arabic data improves the model's representations and robustness for such languages.



\section{Analysis}

\begin{figure}[t!]
\centering
\includegraphics[width=\linewidth]{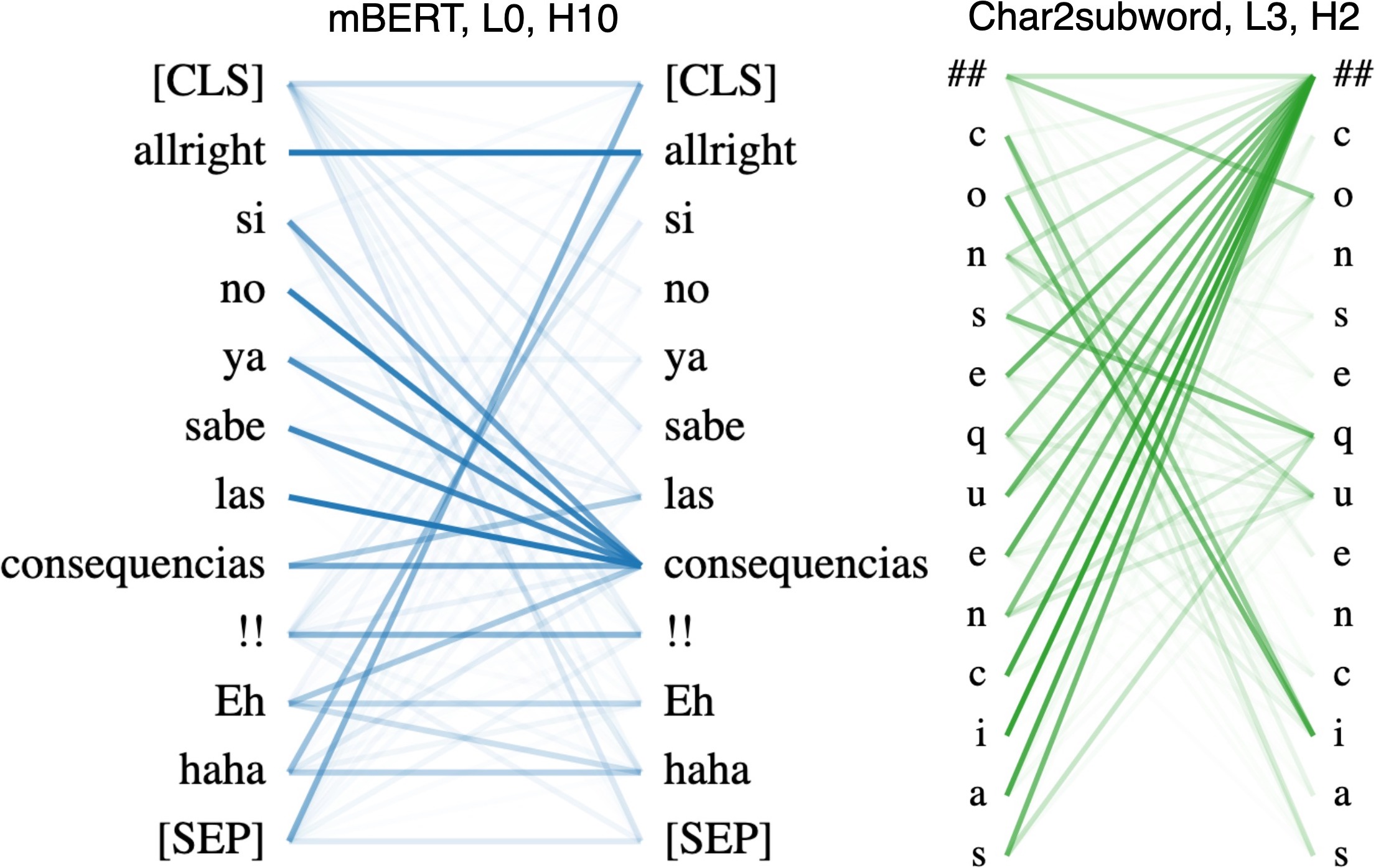}
\caption{
Attention visualization from a Spanish-English tweet. 
Translation: \textit{``Alright, otherwise you know the consequences!! Eh, haha.''} 
}
\label{ch9:fig:csbert_viz_lid}
\end{figure}

\paragraph{Attention for language identification}
Figure \ref{ch9:fig:csbert_viz_lid} shows the visualization for the Spanish-English LID task with an intra-sentential code-switching example (i.e., code-switching at the clause level of a sentence utterance). 
The example shows that the strongest connections at the word level (Figure \ref{ch9:fig:csbert_viz_lid} (left)) happen for words in the same language.
Particularly, the word \textit{consequencias} is slightly ambiguous since its morphology overlaps substantially with both the English and Spanish versions. 
With the context from the surrounding Spanish words, the model can determine that the word is Spanish.
Although there are more patterns captured among all the heads in mBERT, this pattern suggests that words of the same language can provide contextual support along with the sentence.

In addition to the contextual support, character-level attention plays an important role when building the word representation.
Particularly for this word, the ambiguity is introduced due to the letter \textit{q}.
Note that the char2subword module creates strong connections with this letter and parts where more ambiguity could happen.
For example, the letter \textit{i} happens where the suffixes \textit{-cias} (Spanish) and \textit{-ces} English could complete the word).

\paragraph{Error analysis}
By inspecting the mistakes of the model in the confusion matrix for the Spanish-English LID development set, we noticed 112 English words predicted as Spanish, and 101 Spanish words predicted as English (see Table \ref{ch9:fig:cs_bert:confusion_matrix} in Appendix \ref{app:analysis} for the confusion matrix).
Out of the 101 English words, 63 were processed by the char2subword module (i.e., via backoff). 
Most of these errors come from words that heavily overlap in morphology between the two languages. 
For example, the words \textit{imagine, rodeos, superego, tacos} are exact spellings between the languages, while the words \textit{apetite} and \textit{pajamas} change one letter between the languages (e.g., \textit{apetito, pijamas} in Spanish). 
These errors suggest that the robustness may create some ambiguity when it comes to detecting the text's language. 
That is, single-character differences can denote one or another language, but the robustness operations (Table \ref{ch9:tab:noise_operations}) can blur such distinction during the approximation phase.
Other words are interjections that are spelled the same way (e.g., \textit{oh}, \textit{eh}, and \textit{Muahahahahaha}). 
Also, there are cases where the ground-truth labels are wrong. 
E.g., the word \textit{larges} in the the sentence ``\textit{La puerta esta abierta para que te \underline{larges} porque no te has ido}''\footnote{``The door is open for you to leave, why haven't you left?''} was correctly predicted as Spanish based on the context (i.e., the correct spelling is \textit{largues}, which translates to \textit{get out}).

\paragraph{Subword sequence lengths}
Sequences from the subword tokenization are the same length or longer than the original sequence of tokens.
Quantifying that tells us about the opportunity that the char2subword mBERT model has in practice.
Table \ref{ch9:fig:subword_stats_seq_length} shows the statistics of the original sequence lengths (Tokens) and the sequence lengths after the subword tokenization (Subword).
Note that the average sequence lengths tend to duplicate across datasets.
This can potentially explain a larger gap in performance for NER and POS tagging tasks than in LID.
The former tasks require more semantics, which aligns with the fact that subwords degrade meaning by splitting into many pieces.

\begin{table}[t!]
\small
\centering
\resizebox{\linewidth}{!}{
\setlength{\tabcolsep}{3pt}
\begin{tabular}{@{}llrrrrr@{}}
\toprule
& 
& \multicolumn{1}{l}{} 
& \multicolumn{2}{c}{\textbf{Original}}
& \multicolumn{2}{c}{\textbf{Tokenized}} \\ 
\cmidrule(lr){4-5} 
\cmidrule(lr){6-7} 
\textbf{Task} 
& \textbf{Lang.} 
& \textbf{Seqs.}    
& \textbf{Mean\sub{$\pm$Std}} & \textbf{Range}
& \textbf{Mean\sub{$\pm$Std}} & \textbf{Range}
\\
\midrule
\multirow{4}{*}{LID} & es-en   & 3.3K  & 12.1\sub{$\pm$7.7}    & [1, 39]  & 21.1\sub{$\pm$12.0}   & [1, 69]  \\
                     & hi-en   & 744   & 20.8\sub{$\pm$24.1}   & [1, 225] & 31.4\sub{$\pm$32.9}   & [4, 278] \\
                     & ne-en   & 1.3K  & 14.5\sub{$\pm$6.3}    & [3, 34]  & 28.5\sub{$\pm$10.8}   & [3, 63]  \\
                     & msa-arz & 1.1K  & 19.7\sub{$\pm$6.5}    & [2, 36]  & 43.5\sub{$\pm$14.4}   & [2, 93]  \\
\midrule
\multirow{3}{*}{NER} & es-en   & 10K    & 12.1\sub{$\pm$7.6}  & [1, 45]  & 25.7\sub{$\pm$14.2} & [1, 120] \\
                     & hi-en   & 314    & 17.0\sub{$\pm$6.3}  & [4, 34]  & 40.5\sub{$\pm$13.6} & [7, 74]  \\
                     & msa-arz & 1.1K   & 20.2\sub{$\pm$6.7}  & [2, 38]  & 44.5\sub{$\pm$14.8} & [3, 112] \\
\midrule
\multirow{2}{*}{POS} & es-en   & 4.2K  & 7.7\sub{$\pm$6.0}    & [2, 90]  &  9.9\sub{$\pm$7.8}  & [2, 127] \\
                     & hi-en   & 160   & 21.7\sub{$\pm$5.2}   & [5, 37]  & 41.3\sub{$\pm$12.2} & [7, 93]  \\ 
\bottomrule
\end{tabular}
}
\caption{
Statistics across the development sets comparing sequence lengths before (e.g., \textbf{Original}) and after (e.g., \textbf{Tokenized}) subword tokenization.
}
\label{ch9:fig:subword_stats_seq_length}
\end{table}

\paragraph{Parameters vs. efficiency} 
The subword lookup table in mBERT provides immediate access for the tokenized text to the embedding space, making such a table very convenient. 
However, this access is highly restricted to a predefined vocabulary, and, in the case of multilingual models, such vocabulary has to have adequate coverage for all the languages involved. 
Models like mBERT or XLM-R \citep{conneau-etal-2020-unsupervised} use more than 100 languages, which translates into a large number of parameters just to enable the text to be vectorized. 
More specifically, mBERT has 177M parameters in total while only its subword embedding table ($|\mathcal{V}| = 119K$) occupies 91M parameters—more than 50\% of all the parameters of the model.\footnote{For XLM-R base (278M) and large (559M), the percentages are 65\% and 49\%, respectively.}
The char2subword module, on the other hand, reduces the number of parameters to 50M, about 45\% less than the subword embedding table, while also capable of handling misspellings and inflections robustly.
Nevertheless, this module requires more computation time to come up with subword-level embedding representations.

\paragraph{Adversarial attacks} 
We assess the robustness of the char2subword by using the TextAttack library \citep{morris-etal-2020-textattack-framework}.
Particularly, we apply the DeepWordBug recipe \citep{Gao2018BlackBoxGO} to the es-en sentiment analysis validation set.
The attack consists of character-level transformations on the highest-ranked words that minimizes the edit distance of the perturbation.
Notably, the char2subword module is more resilient than mBERT to these attacks; mBERT loses 16.78 points of weighted accuracy (56.10 $\rightarrow$ 39.32), while char2subword + mBERT drops 12.41 points (57.71 $\rightarrow$ 45.30). 
Most of the attacks that affect the prediction on mBERT are entities. Intuitively, this is reasonable since the BPE splits such cases into many subword pieces, while the char2subword sticks to the name words and leverage context.

\section{Conclusion}

We provide a novel, flexible, and robust method to expand the mBERT subword embedding table.
The char2subword module provides more control at the tokenization level, and it can generate word embeddings without being restricted to a fixed vocabulary or segmentation method.
Also, the char2subword module gives the possibility to refine a language or domain of interest (i.e., by pre-training the char2subword module) while preserving its multilingual properties.
Finally, this method is not limited to code-switching; the char2subword module is a general approach that can be applied to any word or subword-based pre-trained model.

\section*{Acknowledgments}
This work was partially funded by the National Science Foundation under grant \#1910192.

\bibliography{anthology,custom}
\bibliographystyle{acl_natbib}

\appendix

\section{Char2subword Module Definition} 
\label{app:char2subword_definition}

We model the char2subword module $f_\theta$ using the Transformer architecture \citep{vaswani2017attention}.
The module processes a sample as a sequence of characters $\vc_i = (c_{i1}, c_{i2}, \dots, c_{iM})$ 
of a subword $s_i$ of length $M$.\footnote{
To distinguish between words and subwords, we prepend `\#\#' to the sequence $\vc_i$ in the case of full words.
}
We represent the sequence $\vc_i$ as the sum between the character embeddings and sinusoidal positional encodings.
We pass the resulting sequence of character vectors $\mX_0$ to a stack of $l$ attention layers, each with $k$ attention heads. 
The $j$-th attention layer receives the input $\mX_j$ and it outputs $\mX_{j+1}$ by applying two subsequent components: multi-head attention and feed-forward layers. 
The multi-head attention is defined as follows:
\resizebox{\hsize}{!}{
\begin{minipage}{\hsize}
\begin{align*}
    \mathrm{Attn}(\mQ, \mK, \mV) &= \mathrm{softmax}(\frac{\mQ\mK^\top}{\sqrt{d'}}) \mV \\
    \mathrm{MultiHead}(\mX) &= [\mathrm{head}_1; \dots; \mathrm{head}_k]\mW^O \\
    \mathrm{where}~ \mathrm{head}_i &= \mathrm{Attn}(\mX \mW^{Q_i}_{j}, \mX \mW^{K_i}_{j}, \mX \mW^{V_i}_{j}) \\
    \mX'_j &= \mathrm{MultiHead}(\mX_j)
\end{align*}
\vspace{0.01\baselineskip}
\end{minipage}
}

The feed-forward component linearly projects $\mX'_j$ using $\mW_{j1} \in \sR^{d'\times4d'}$ followed by a GELU activation function \citep{hendrycks-and-gimple-2016-gelu}.
The projection is passed to another linear transformation such that the result $\mX'_j$ is mapped back to $\sR^{d'}$:
$$
\mathrm{FFN}(\mX'_j) = \mathrm{GELU}(\mX'_j \mW_{j1} + \vb_{j1})\mW_{j2} + \vb_{j2}
$$
Each component normalizes its input $\bar{\mX}_j = \mathrm{LayerNorm}(\mX_j)$ using layer normalization \citep{ba2016layer}. 
We add the normalized input to the output of the component as in a residual connection \citep{he-etal-2016-residual}:
\begin{align*}
    \mX'_j &= \mathrm{MultiHead}(\bar{\mX}_j) + \bar{\mX}_j \\
    \mX_{j+1} &= \mathrm{FFN}(\bar{\mX'}_j) + \bar{\mX'}_j
\end{align*}

Following \cite{vaswani2017attention}, we preserve the dimension $d'$ of the character embedding throughout the attention layers. On top of the $l$ attention layers, we add a linear layer $\mW_e \in \sR^{d' \times d}$ followed by max-pooling and a layer normalization for the final output $\hat{\ve}_i$:
$$
\hat{\ve}_i = \mathrm{LayerNorm}(\mathrm{maxpool}(\mX_l \mW_e + \vb_e))
$$

\section{Analysis}
\label{app:analysis}

In Table \ref{ch9:fig:cs_bert:confusion_matrix}, we provide the confusion matrix of the pre-trained char2subword model on the Spanish-English LID development set. 

\begin{table}[t!]
\small
\centering
\resizebox{\linewidth}{!}{
\setlength{\tabcolsep}{3pt}
\begin{tabular}{lrrrrrrrrr}
\toprule
& \multicolumn{8}{c}{\textbf{Ground-truth}}\\\cmidrule(lr){2-9}
\textbf{Pred.}
& \textbf{amb.} 
& \textbf{fw} 
& \textbf{lang1} 
& \textbf{lang2} 
& \textbf{mixed} 
& \textbf{ne} 
& \textbf{other} 
& \textbf{unk} 
\\
\midrule
\textbf{amb.}   & 0     & 0    & 21       & 16       & 0    & 0      & 1       & 1   \\
\textbf{fw}          & 0     & 1    & 0        & 1        & 0    & 0      & 0       & 0   \\
\textbf{lang1}       & 14    & 0    & 16K   & 101      & 0    & 74     & 14      & 17  \\
\textbf{lang2}       & 13    & 0    & 112      & 14K   & 0    & 51     & 5       & 3   \\
\textbf{mixed}       & 0     & 0    & 1        & 4        & 0    & 1      & 0       & 0   \\
\textbf{ne}          & 3     & 0    & 110      & 96       & 1    & 597    & 7       & 1   \\
\textbf{other}       & 1     & 0    & 13       & 6        & 1    & 3      & 7K   & 4   \\
\textbf{unk}         & 0     & 0    & 8        & 10       & 0    & 3      & 3       & 8   \\
\bottomrule
\end{tabular}
}
\caption{
    The confusion matrix on the development set of the LID task for Spanish-English. 
    The labels are lang1 (English), lang2 (Spanish), mixed (partially in both languages), ambiguous (either one or the other language), fw (a language different than lang1 and lang2), ne (named entities), other, and unk (unrecognizable words). 
}
\label{ch9:fig:cs_bert:confusion_matrix}
\end{table}

\end{document}